\pdfoutput=1
\RequirePackage{amsmath}
\RequirePackage{amssymb}
\documentclass[runningheads]{llncs}
\usepackage{tabulary}
\usepackage{graphicx}
\usepackage{comment}
\usepackage{bbm}
\usepackage[ruled,vlined]{algorithm2e}

\usepackage[title]{appendix}
\newcommand{\repeatthanks}{\textsuperscript{\thefootnote}}

\begin{document}
\title{Salient Facial Features from Humans and Deep Neural Networks}
\titlerunning{Salient Facial Features from Humans and Deep Neural Networks}


%
 \author{Shanmeng Sun\inst{1}\thanks{The authors contributed equally} \and
 Wei Zhen Teoh\inst{2,3}\repeatthanks
 \and
 Michael Guerzhoy\inst{4,5,6}}
 \authorrunning{S. Sun et al.}
%
\institute{Affiliation\
\\
\email{email}\
}
 
 \institute{
 Advanced Micro Devices, Markham, Ontario, Canada
 \\
 \email{shanmeng.sun@amd.com,  wezteoh@descript.com, guerzhoy@princeton.edu}
 \and
 Lyrebird AI, Montreal, Québec, Canada
 \and
 Descript, San Francisco, California, USA
 \and 
 University of Toronto, Toronto, Ontario, Canada  
 \and
 Li Ka Shing Knowledge Institute, Toronto, Ontario, Canada
 \and
 Princeton University, Princeton, New Jersey, USA
 }

\maketitle              

\begin{abstract}
 In this work, we explore the features that are used by humans and by convolutional neural networks (ConvNets) to classify faces. We use Guided Backpropagation (GB) to visualize the facial features that influence the output of a ConvNet the most when identifying specific individuals; we explore how to best use GB for that purpose.  We use a human intelligence task to find out which facial features humans find to be the most important for identifying specific individuals. We explore the differences between the saliency information gathered from humans and from ConvNets.
 
 Humans develop biases in employing available information on facial features to discriminate across faces. Studies show these biases are influenced both by neurological development and by each individual's social experience. In recent years the computer vision community has achieved human-level performance in many face processing tasks with deep neural network-based models. These face processing systems are also subject to systematic biases due to model architectural choices and training data distribution. 

\end{abstract}

\section{Introduction}
Recognizing faces is important in human society. Faces convey a wealth of information about people, including identity, age, gender, and race. Studies~\cite{LOGAN201729}~\cite {facearticle} have shown that humans develop spatial biases in accessing information available on facial features to discriminate between faces. Face processing has been shown to also be influenced by individual social experiences~\cite{10.1093/scan/nsm035}. In this work we are interested in examining biases in identifying faces through visualization. In literature, the term \textit{saliency} has been coined to refer to the level by which a spatial region stands out visually from its neighbourhood. In our context, we use \textit{facial feature saliency} as the level to which a particular feature makes the face identifiable or distinguishable by a decision agent. Traditional studies to examine facial feature saliency from humans involve eye-tracking experiments to gather the required data~\cite{7410802}. In this work, we develop a task to collect such data from humans by requiring them to draw bounding boxes around regions they think is salient on facial images.

There is reason to think that deep learning vision models have different biases than humans. For instance, deep convolutional neural network (ConvNet) models exhibit biases towards texture rather than shape in visual classification task~\cite{DBLP:journals/corr/abs-1811-12231}. In face recognition and classification tasks, we are interested in  finding out which facial features are important for the task. We use the \textit{Guided Backpropagation (GB)} technique developed in ~\cite{springenberg2014striving} to study facial saliency derived from a ConvNet model that we trained on facial images. GB assigns a saliency to each pixel in the input.

In Section~\ref{classifier}, we introduce our ConvNet-based classifier, and demonstrate how Guided Backpropagation can be used to visualize features that are important for the ConvNet's classifying faces as belonging to a particular individual. In Section~\ref{mturk}, we describe a human intelligence task that elicits human opinions of which features are salient. We compare the features that are important to ConvNets to the features that are identified as important to humans in Section~\ref{compare}.

\section{Deep ConvNet Classifier-Derived Saliency \label{classifier}}

\subsection{Face Classifier Model}
We train a face classifier to predict the identity of the person in a face image. We obtained 100 $227\times 227$ images for each of the 17 selected actors (8 male, 9 female) from the FaceScrub dataset~\cite{7025068}. Each image is cropped to a tight square around the face. 15 images of each actor are held out as test set.

\begin{figure}
\centering
  \includegraphics[width=0.85\linewidth]{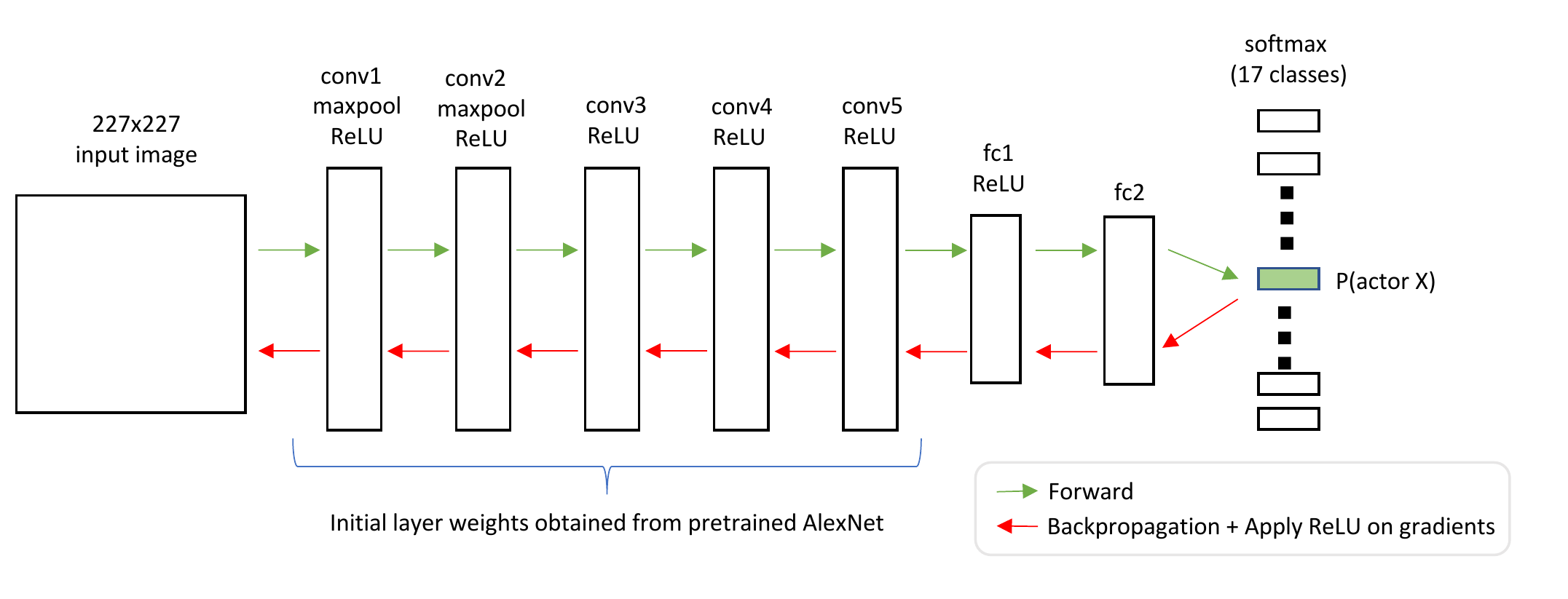}
\vspace{-5mm}
\caption{Augmented AlexNet model architecture trained as a face classifier. Starting from a target class probability output, the guided backpropagation is performed by backpropagating normally from layer to layer, and zeroing out the the negative gradients after each backpropagation step.}
\label{fig:alexnet}
\end{figure}

As a starting point, we use the AlexNet~\cite{NIPS2012_4824} model pre-trained on the ImageNet~\cite{ILSVRC15} dataset for transfer learning. All the convolutional layer weights of the network are retained and we connect the outputs of the last convolutional layer of AlexNet to two fully connected layer. The new hidden layer is of size 100; the output layer is of size 17 (the number of outputs), and is passed through a Softmax. The augmented model is shown in Fig.\ref{fig:alexnet}.  We train two variants of the classifier: (a) we hold the pre-trained weights constant and train only the newly added layers to the truncated network; and (b) we subsequently fine-tune the whole network end-to-end. The loss function is a cross-entropy formulation with weight decay regularization across all layers.

\subsection{Guided Backpropagation (GB)}
For each test image, we compute a visual explanation for the correct output of the model on the image by using Guided Backpropagation (GB)~\cite{springenberg2014striving}. GB is computed using a modification on the process of computing the gradient of the output of interest (e.g. the output neuron correspond to individual A) with respect to the input image. Operationally, it means that we perform Backpropagation from layer to layer and zero out the negative gradients after each step. In our context, the GB gradient of a class $y$ with respect to image $X$ across $N$ intermediate layers can be computed using the following algorithm:

\begin{algorithm}[H]
\SetStartEndCondition{}{}{}%
\SetKwProg{Fn}{def}{\string:}{}
\SetKwFunction{Range}{range}
\SetKw{KwTo}{in}\SetKwFor{For}{for}{\string:}{}%
\SetKwIF{If}{ElseIf}{Else}{if}{:}{elif}{else:}{}%
\SetKwFor{While}{while}{:}{fintq}%
\AlgoDontDisplayBlockMarkers\SetAlgoNoEnd\SetAlgoNoLine%
\SetKwFunction{fmatmul}{matmul}
\SetKwFunction{frelu}{relu}
\SetKwFunction{fgb}{GB}

\Fn{\fgb{$y$, $X$}}{
   grad $\gets$ \frelu($dP(y)/dL_N$)\;
   \For{$n \gets N-1\ to\ 1$}{
    grad $\gets$ \frelu(\fmatmul(grad, $dL_{n+1}/dL_{n}$))\;
   }
   grad $\gets$ \frelu(\fmatmul(grad, $dL_1/dX$))\;
 \Return grad}
 \caption{Guided Backpropagation}
\end{algorithm}

\begin{figure}
\centering
  \includegraphics[width=\linewidth]{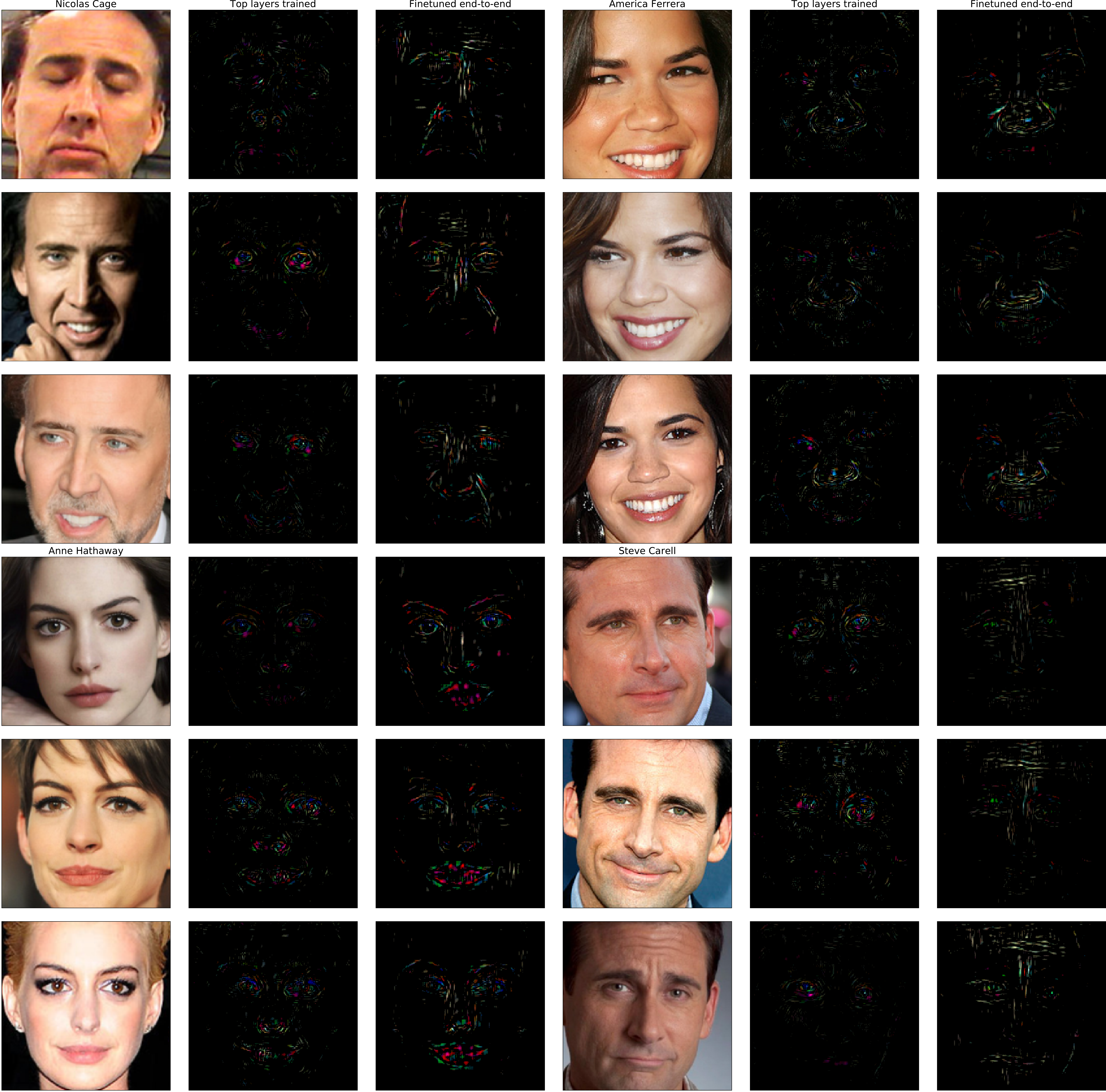}
\vspace{0mm}
\caption{More features are retained when displaying the GB visulization of the network trained end-to-end. We can see laugh lines of Nicolas Cage, lips of Anne Hathaway, nose of America Ferrera and forehead wrinkles of Steve Carell.}
\label{fig:tl_vs_e2e}    
\end{figure}

We can interpret GB gradient values as an indicator of the level to which the pixels in the input image make the face identifiable as a particular individual. We introduce the term \textbf{classifier saliency maps} to refer to the map of GB gradient values on the facial image. 
In our experiments, GB produces different-looking saliency maps for different individuals when applied to networks that were fine-tuned end-to-end. More edges are retained when visualizing the classifier trained end-to-end. See Fig. \ref{fig:tl_vs_e2e}. In these saliency map visualizations we mask out all but the top 5 percent of pixels by intensity. A large portion of the pretrained convolutional layer weights of AlexNet go to 0 during end-to-end fine-tuning with weight decay.

We also observe that saliency maps generated by GB are consistent across different photos of the same individual. We compute the $R^2$ value and find that 92\% of the variation in the GB visualizations is explained by the class of the images. (This is no doubt partly due to the fact that images of the same individual look like each other; however, note that we only retain 5\% of the pixels.)

\subsubsection{Saliency Differences} GB can be used to answer the question  ``how is the facial appearance of individual $y$ different from individual $z$?.'' Using an image X from individual $y$, we compute GB gradients of class $y$ and GB gradients of class $z$ with respect to image $X$. These saliency maps highlight the facial regions that make the image more likely to be classified as class $y$ and class $z$ respectively. We compute the saliency difference between class $y$ and $z$ of image $X$ using

\begin{equation}
Saliency\ difference_{y-z}(X) =  ReLU(GB(y, X) - GB(z, X))
\end{equation}

\begin{figure}[ht]
  \includegraphics[width=\linewidth]{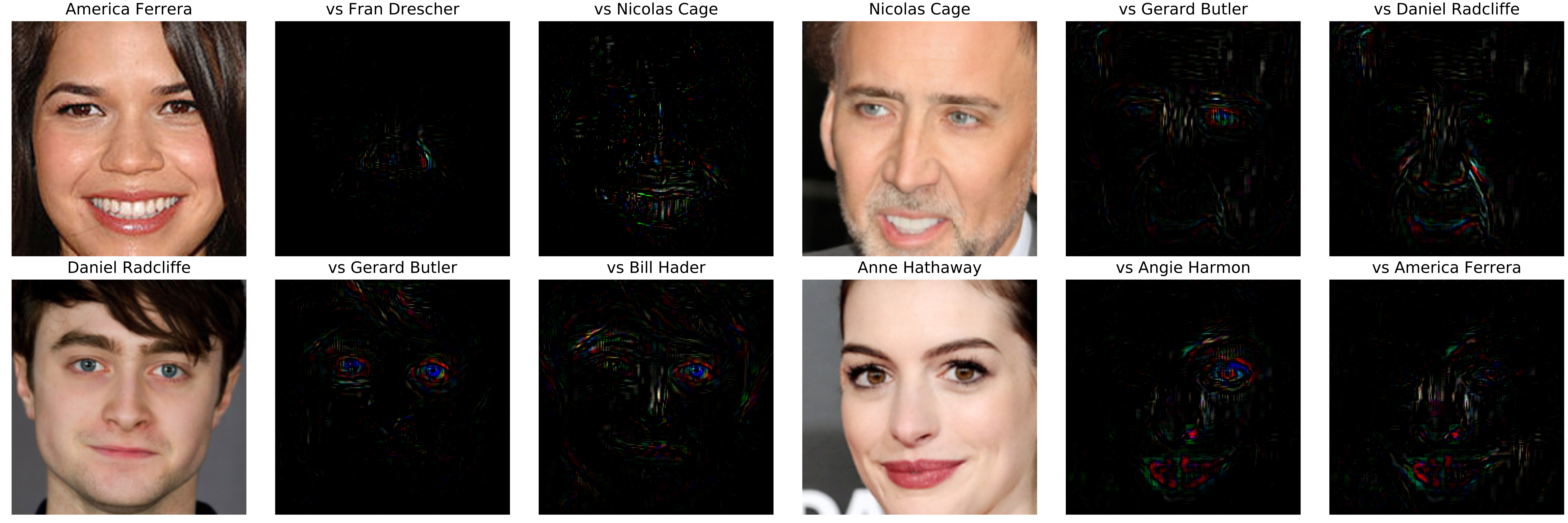}
  \caption{We compute saliency differences between the correct class and two incorrect classes. America Ferrera's nose appears to distinguish her from Fran Drescher and her teeth pattern distinguishes her from Nicolas Cage. Nicolas Cage's eyes distinguish him from Gerard Butler and his laugh lines distinguish him from Daniel Radcliffe. Daniel Radcliffe's blue eyes and Anne Hathaway's deep red lips consistently set them apart from the others.}
  \label{fig:pairwise}
\vspace{-5.5mm}
\end{figure}

We visualize the saliency differences using the test images in Fig.~\ref{fig:pairwise}. 

\section{Human-identified Salient Facial Features}
 
\subsection{Crowdsourced Data from Human Intelligence Task\label{mturk}}
We aim to find which facial features humans typically rely on to distinguish  different faces. We developed a human intelligence task which is deployed on the Amazon Mechanical Turk (MT) platform to obtain human perception data. For this task we used the same test images mentioned in section~\ref{classifier}. A human worker is presented with a randomly selected image from the image pool. The worker is then asked to draw a bounding box around the most recognizable and differentiating facial feature for this individual and label the selected region. A list of 11 labels was provided; however, the workers can also create their own labels. The labels provided are: beard, cheek, chin, ears, eyes, eye brows, hairline, laugh line, lips, moustache, and nose. We collected a total of 5354 responses across all 17 individuals.

\begin{figure}
\centering
  \includegraphics[width=\linewidth]{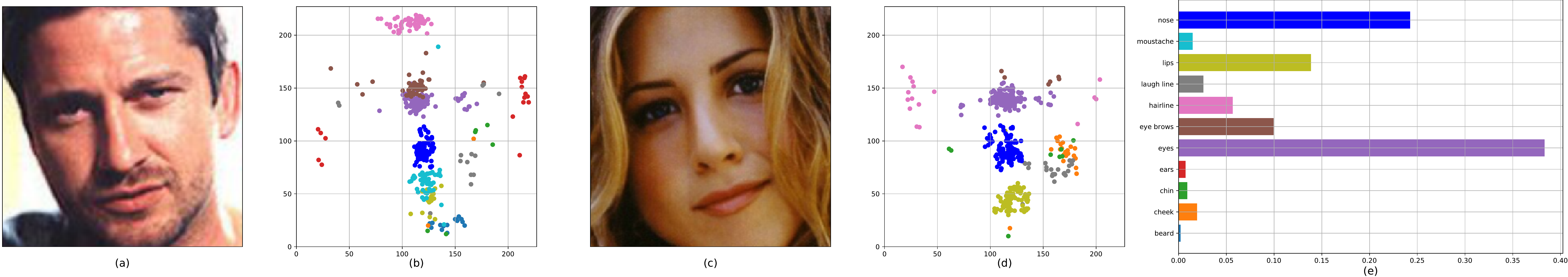}
\vspace{-5.5mm}
\caption{a) sample image of Gerard Butler, b) center of bounding boxes drawn by human workers across for all 15 images of Gerard Butler, c) sample image of Jennifer Aniston, d) center of bounding boxes for Jennifer Aniston, e) normalized histogram of labels tagged by human workers (we balance the contributions from each individual's images  to prevent the difference in response counts from skewing the results).}
\label{fig:faceplot}    
\end{figure}

\section{Deep ConvNet Classifier Derived Saliency vs Human Identified Saliency \label{compare}}

\subsection{Visualizing the ``Biases"}
\begin{figure}
\centering
  \includegraphics[width=0.95\linewidth]{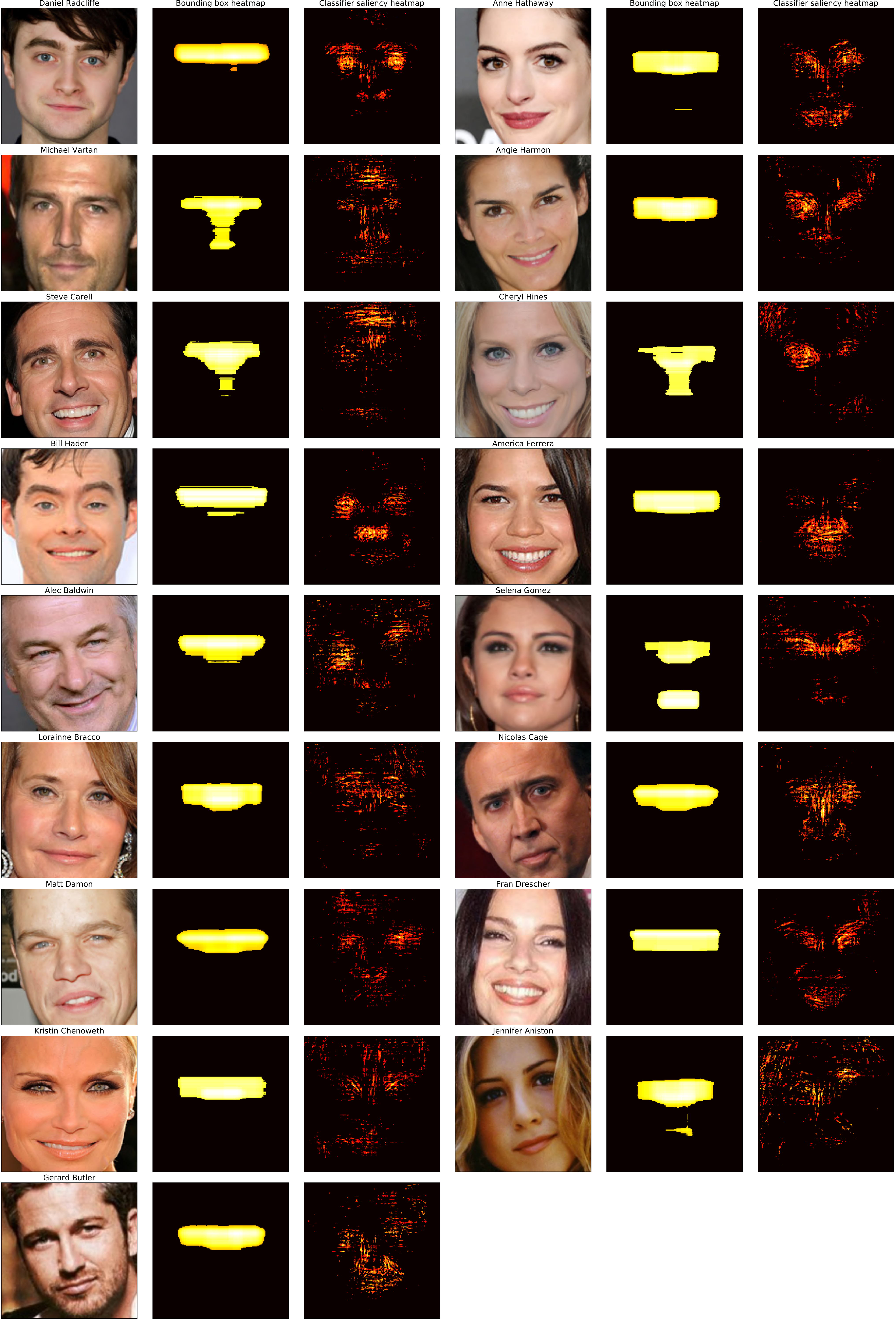}
\vspace{0mm}
\caption{Human workers' results exhibit a strong bias for the region around the eyes, while the classifier picks up information that is more diverse. GB on classifier highlights influence of edge information such as eye wrinkles and forehead wrinkles.}
\label{fig:heatmap}    
\end{figure}

For each individual, we generate a bounding box heatmap and a classifier saliency heatmap. See  Fig. ~\ref{fig:heatmap}. The bounding box heatmap shows how frequent each pixel point is located inside the bounding box drawn by the human workers. We only highlights the pixel locations above the 90th percentile in the visualization as it better localizes the most selected facial region.

The classifier saliency heatmaps aggregate information across saliency maps on all test images for each individual. We first generate a saliency map for each of an individual's 15 test images by applying GB with respect to the correct output. We apply an indicator function to only preserve the top 5 percent pixel by intensity. We average across the filtered saliency maps from all the individual's images to yield classifier saliency heatmap. We only highlight $5\%$ of the pixel locations with the highest average value in our visualizations.

As shown in Fig. ~\ref{fig:heatmap}, human workers consistently pick out the same local region around the eyes as the most salient facial feature for each individual. There is a lot less variation in features picked out by human workers as compared the the Deep ConvNet classifier. While the eyes are also commonly picked out by GB, the classifier shows tendency to use edge and texture information in distinguishing across faces. For example, the forehead wrinkles are clearly highlighted for some of the male individuals. It is also noticeable that the deep classifier exhibits less symmetry in accessing information. For instance, while human workers tend to highlight both eyes, the classifier clearly highlights one eye significanntly more than the other for Cheryl Hines. 

\begin{figure}
\centering
  \includegraphics[width=0.95\linewidth]{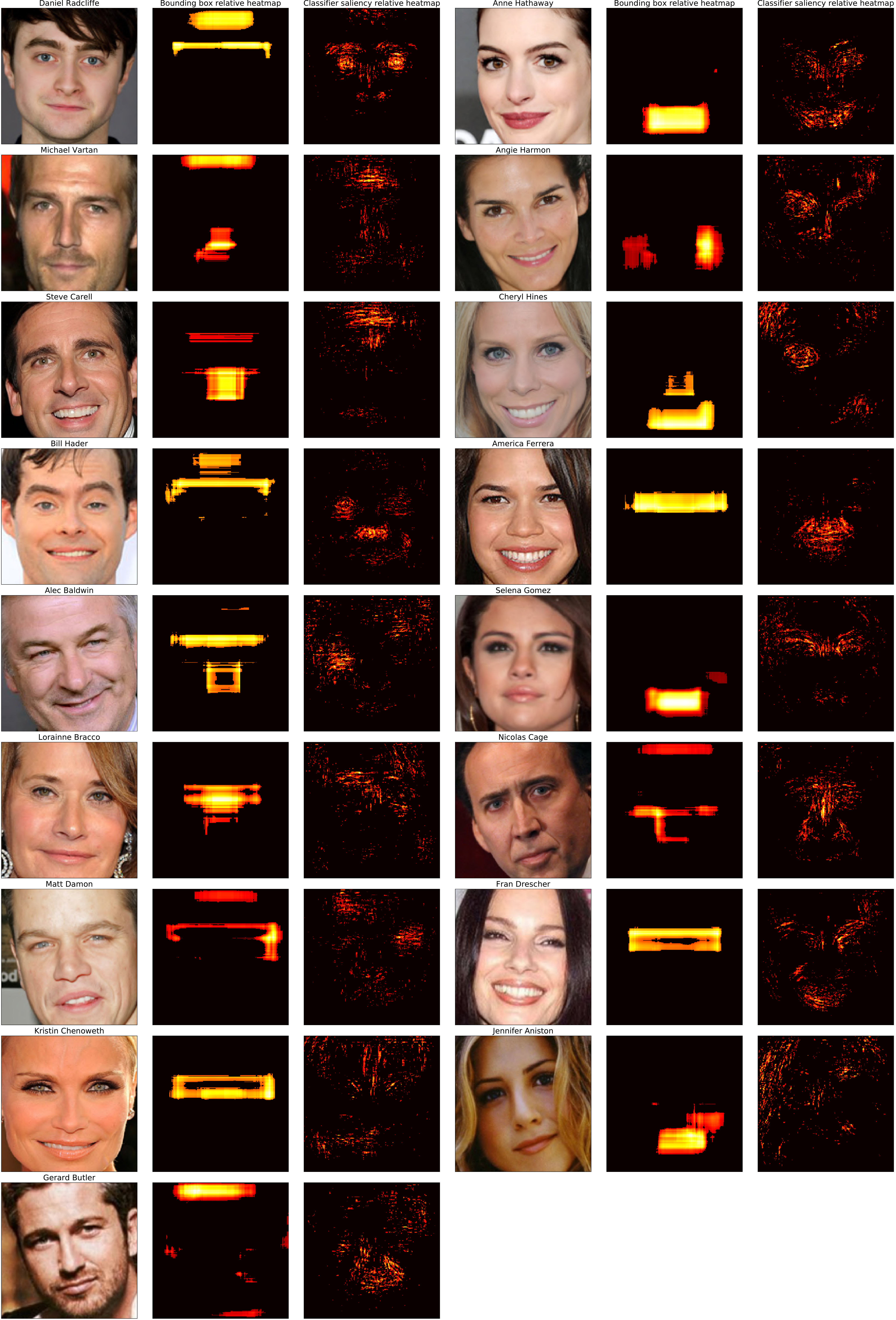}
\vspace{0mm}
\caption{}
\label{fig:relative_heatmap}    
\end{figure}

\subsubsection{Relative Saliency} 
While human workers appear to heavily prefer certain facial feature over others, we can still observe which facial feature of a person they find relatively special compared with others. We obtain an average bounding box heatmap by averaging across the bounding box heatmaps of all individuals. By taking the difference between the individual's bounding box heatmap and the average heatmap, we produce a relative saliency visualization. By similar approach we generate saliency relative heatmaps from the classifier. We only highlight the top 10\% and 5\% pixels by intensity in the bounding box heatmaps from humans and saliency heatmaps from classifier respectively in our visualizations (Fig. ~\ref{fig:relative_heatmap}). Even at relative basis, there appears to be no consistent correspondence observed between saliency information between GB and human workers.

\section{Limitations}
In this work, we have only explored the subject of facial feature saliency based on a small, racially homogeneous dataset. The classifier trained on this dataset only optimizes for recognizing intra-dataset differences across specific faces, while human workers are ``trained" based on their more diverse personal experiences. In addition, in the human intelligence task, some labels were provided as a guidance which could arguably bias the responses. These differences in the scale of facial comparison could explain the differences observed.

Human workers were asked to pick rectangles that corresponded to distinguishing facial features, while Guided Backpropagation is a modification of the gradient of the correct output with respect to the input. Those tasks are analoguous but not equivalent.

\section{Conclusion and Future work}
In this work, we investigated the differences between facial feature saliency identified by humans and by a deep neural network. We explored the approach of Guided Backpropagation to obtain spatial information on facial feature saliency from a trained ConvNet model. We developed a human intelligence task that allowed us to crowdsource data on localization of facial regions identified as salient by humans. Lastly, we created visualizations that suggest differences in face processing between humans and a deep neural network.

A topic worth exploring further is whether different classifier models exhibit similar biases. The variables in play could include architectural choices, the dataset that the model is pretrained on and model capacity. Another fundamental question to be answered on human perception is whether human workers correctly identify regions that actually are the most salient to them: how do voluntary data from humans compare against outputs from an eye-tracking experiment? The impact of facial memory is also unclear to us, i.e. whether workers' familiarity with the faces influences the results. Answers to these questions will inform improvement in experimental design for more detailed assessments.


%
%
%
\clearpage
\bibliographystyle{splncs04}
\bibliography{bib}
\clearpage

\end{document}